\documentclass{article}

\usepackage{arxiv}

\usepackage[utf8]{inputenc} % allow utf-8 input
\usepackage[T1]{fontenc}    % use 8-bit T1 fonts
\usepackage{hyperref}       % hyperlinks
\usepackage{url}            % simple URL typesetting
\usepackage{booktabs}       % professional-quality tables
\usepackage{amsfonts}       % blackboard math symbols
\usepackage{nicefrac}       % compact symbols for 1/2, etc.
\usepackage{microtype}      % microtypography
\usepackage{lipsum}         % Can be removed after putting your text content
\usepackage{graphicx}
\usepackage{natbib}
\usepackage{doi}

\usepackage{subfig}
\usepackage{float}
\usepackage{lineno}
\usepackage{amsmath}
\usepackage{tikz}
\usepackage{cleveref}       % smart cross-referencing

\title{RocSync: Millisecond-Accurate Temporal Synchronization for Heterogeneous Camera Systems}

\usepackage{authblk}

\newbox{\orcid}\sbox{\orcid}{\includegraphics[scale=0.06]{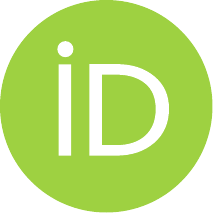}} 
\author[1]{%
	\href{https://orcid.org/0009-0008-6408-5978}{\usebox{\orcid}\hspace{1mm}Jaro Meyer}%
}
\author[2]{%
	\href{https://orcid.org/0000-0001-6484-6206}{\usebox{\orcid}\hspace{1mm}Frédéric Giraud}%
}
\author[1]{%
	\href{https://orcid.org/0009-0000-8886-3182}{\usebox{\orcid}\hspace{1mm}Joschua Wüthrich}%
}
\author[1]{%
	\href{https://orcid.org/0000-0003-2448-2318}{\usebox{\orcid}\hspace{1mm}Marc Pollefeys}%
}
\author[2]{%
	\href{https://orcid.org/0000-0001-6484-6206}{\usebox{\orcid}\hspace{1mm}Philipp Fürnstahl}%
}
\author[2]{%
	\href{https://orcid.org/0000-0002-2565-2297}{\usebox{\orcid}\hspace{1mm}Lilian Calvet\thanks{Correspondence: \url{Lilian.Calvet@balgrist.ch}}}%
}
\affil[1]{ETH Zurich, Zurich, Switzerland}
\affil[2]{Research in Orthopedic Computer Science, Balgrist University Hospital, University of Zurich, Zurich, Switzerland}

\renewcommand{\shorttitle}{RocSync}

\hypersetup{
pdftitle={RocSync: Millisecond-Accurate Temporal Synchronization for Heterogeneous Camera Systems},
pdfauthor={Jaro Meyer, Frédéric Giraud, Joschua Wüthrich, Marc Pollefeys, Philipp Fürnstahl, Lilian Calvet},
pdfkeywords={video synchronization; heterogeneous camera systems; sub-frame temporal alignment; multi-view 3D reconstruction; multi-view pose estimation; infrared and RGB imaging; computer vision},
}

\begin{document}
\maketitle

\begin{abstract}
	Accurate spatiotemporal alignment of multi-view video streams is essential for a wide range of dynamic-scene applications such as multi-view 3D reconstruction, pose estimation, and scene understanding. However, synchronizing multiple cameras remains a significant challenge, especially in heterogeneous setups combining professional and consumer-grade devices, visible and infrared sensors, or systems with and without audio, where common hardware synchronization capabilities are often unavailable. This limitation is particularly evident in real-world environments, where controlled capture conditions are not feasible. In this work, we present a low-cost, general-purpose synchronization method that achieves millisecond-level temporal alignment across diverse camera systems while supporting both visible (RGB) and infrared (IR) modalities. The proposed solution employs a custom-built \textit{LED Clock} that encodes time through red and infrared LEDs, allowing visual decoding of the exposure window (start and end times) from recorded frames for millisecond-level synchronization. We benchmark our method against hardware synchronization and achieve a residual error of 1.34~ms RMSE across multiple recordings. In further experiments, our method outperforms light-, audio-, and timecode-based synchronization approaches and directly improves downstream computer vision tasks, including multi-view pose estimation and 3D reconstruction. Finally, we validate the system in large-scale surgical recordings involving over 25 heterogeneous cameras spanning both IR and RGB modalities. This solution simplifies and streamlines the synchronization pipeline and expands access to advanced vision-based sensing in unconstrained environments, including industrial and clinical applications.
\end{abstract}

% keywords

% sections
\makeatletter
\renewcommand\paragraph[1]{%
  \vspace{1ex}% optional small space before (remove if not wanted)
  \noindent\textbf{#1. }\ignorespaces
}
\makeatother

% Defining the magnifying glass macro
\newcommand{\magnifyimage}[6]{
    % #1: image path, #2: x-center, #3: y-center, #4: radius
    \begin{tikzpicture}
        % Load the original image (full, unclipped)
        \node[inner sep=0] at (0,0) {\includegraphics[width=5cm]{#1}};

        % Define parameters
        \def\magcenterx{#2} % x-coordinate of the magnifying glass center
        \def\magcentery{#3} % y-coordinate of the magnifying glass center
        \def\magradius{#4}  % Radius of the magnifying glass circle
        \def\magscale{2.1}  % Magnification factor
        \def\maginsetx{#5}   % x-offset for magnified portion (bottom-right)
        \def\maginsety{#6}  % y-offset for magnified portion (bottom-right)

        % Draw the magnified portion in the bottom-right corner
        \begin{scope}
            % Clip at the correct position in the original image's coordinate system
            \clip (\maginsetx, \maginsety) circle (\magradius*\magscale);
            % Place the scaled image so the magnified center aligns with the clip
            \node[inner sep=0, scale=\magscale] at 
                (\maginsetx - \magcenterx*\magscale, \maginsety - \magcentery*\magscale) 
                {\includegraphics[width=5cm]{#1}};
        \end{scope}

        % Draw a circle to indicate the magnified area on the original image
        \draw[thick, red] (\magcenterx,\magcentery) circle (\magradius);

        % Draw a border around the magnified portion to simulate a lens
        \draw[thick, red] (\maginsetx,\maginsety) circle (\magradius*\magscale);
    \end{tikzpicture}
}

\section{Introduction}\label{sec:intro}

% explaining the problem
Temporal synchronization of multi-view video streams is essential for a wide range of computer vision tasks, including 3D reconstruction~\cite{mildenhall2020nerfrepresentingscenesneural, kerbl3Dgaussians, wang2024learningbasedmultiviewstereosurvey}, pose estimation~\cite{9157049, 9758679, HEIN2025103613}, and scene understanding~\cite{10611736, 10.5555/3666122.3667810, rodin2024action, li2025egosplatopenvocabularyegocentricscene}. The exact tolerance to synchronization errors depends heavily on the application. Semantic scene analysis tasks, such as surgical phase recognition, activity classification, or object detection, tend to aggregate information over extended temporal windows, making them relatively robust to minor temporal misalignments. In contrast, geometry-driven applications like multi-view 3D reconstruction and motion capture require that corresponding frames represent the same physical moment. In these cases, even single-digit millisecond offsets can cause significant reprojection errors, blurry artifacts in reconstructions, or inaccurate motion trajectories.

% existing methods, and why they dont always work
In laboratory environments, specialized cameras with dedicated interfaces for hardware-based synchronization~\cite{azure_kinect_multi_sync, basler_ace2, emergent_hr_series} can be used to ensure that the shutters are perfectly synchronized. However, such solutions are costly, require complex wired setups, and greatly restrict the range of compatible devices. In contrast, real-world deployments often involve camera arrays that integrate a heterogeneous mix of consumer and professional devices, many of which lack hardware synchronization capabilities.

When hardware synchronization is not available, temporal alignment is often approximated by recording unsynchronized videos and then, during post-processing, estimating time offsets and pairing each frame from one camera with the closest frame from another stream.

In practice, audio-based methods, such as clapping or other transient sound cues, remain common due to their simplicity and widespread availability of built-in microphones~\cite{ozsoy2025mmorlargemultimodaloperating, ozsoy2025egoexoregoexocentricoperatingroom, NEURIPS2024_c1017d0a}. However, these methods are sensitive to environmental noise and cannot be used with devices lacking audio capture. Moreover, they constrain the spatial arrangement of the cameras, as all cameras must be within audible range and preferably positioned at similar distances from the sound source; otherwise, the relatively low propagation speed of sound causes arrival-time differences that can compromise synchronization. For example, in a room measuring 10 meters across, the sound of a clap will take roughly 30 milliseconds to travel from one side to the other, given the speed of sound in air ($\approx 343$~m/s). This delay is already on the order of an entire video frame at 30~fps.

One alternative solution is to use timecodes stored alongside the video~\cite{GoPro_HERO12_TimecodeSync, li2022neural3dvideosynthesis}, but this requires external timecode systems or pre-synchronizing the cameras’ internal clocks before recording. More recently, content-based approaches have been proposed that estimate temporal relationships directly from image content, for example by tracking general scene features~\cite{shin2025audioposegeneralpurposeframework}, motion patterns~\cite{10.1007/978-3-642-32717-9_27}, lighting changes~\cite{smid2017rolling}, or human poses~\cite{9022391} across multiple cameras. While some of these methods achieve millisecond-level synchronization accuracy, they are typically tailored to narrow use cases and rely on specific video content, which limits their applicability in more diverse scenarios. Some systems address this by explicitly embedding visually encoded timecodes into the video~\cite{s24216975}. However, existing approaches are constrained by the low and fixed refresh rates of consumer screens and are not suitable for scenarios involving infrared (IR) cameras.

In contrast to hardware synchronization, post hoc alignment methods cannot ensure perfect frame correspondence, since the camera shutters operate independently. The residual error is bounded by half the frame period and thus scales with frame rate. For example, at 30~fps, the maximum misalignment can reach 16.7~ms, which is significant for fast motion. Although higher frame rates reduce this error, they also increase storage, power consumption (especially critical for battery-powered cameras), bandwidth, and processing costs, which can be prohibitive for long video sequences such as four-hour surgical recordings. It is therefore beneficial to estimate time offsets with sub-frame precision, enabling accurate frame interpolation for many downstream tasks.

\begin{figure}[t]
\centering
\subfloat[\centering]{\includegraphics[height=3.3cm]{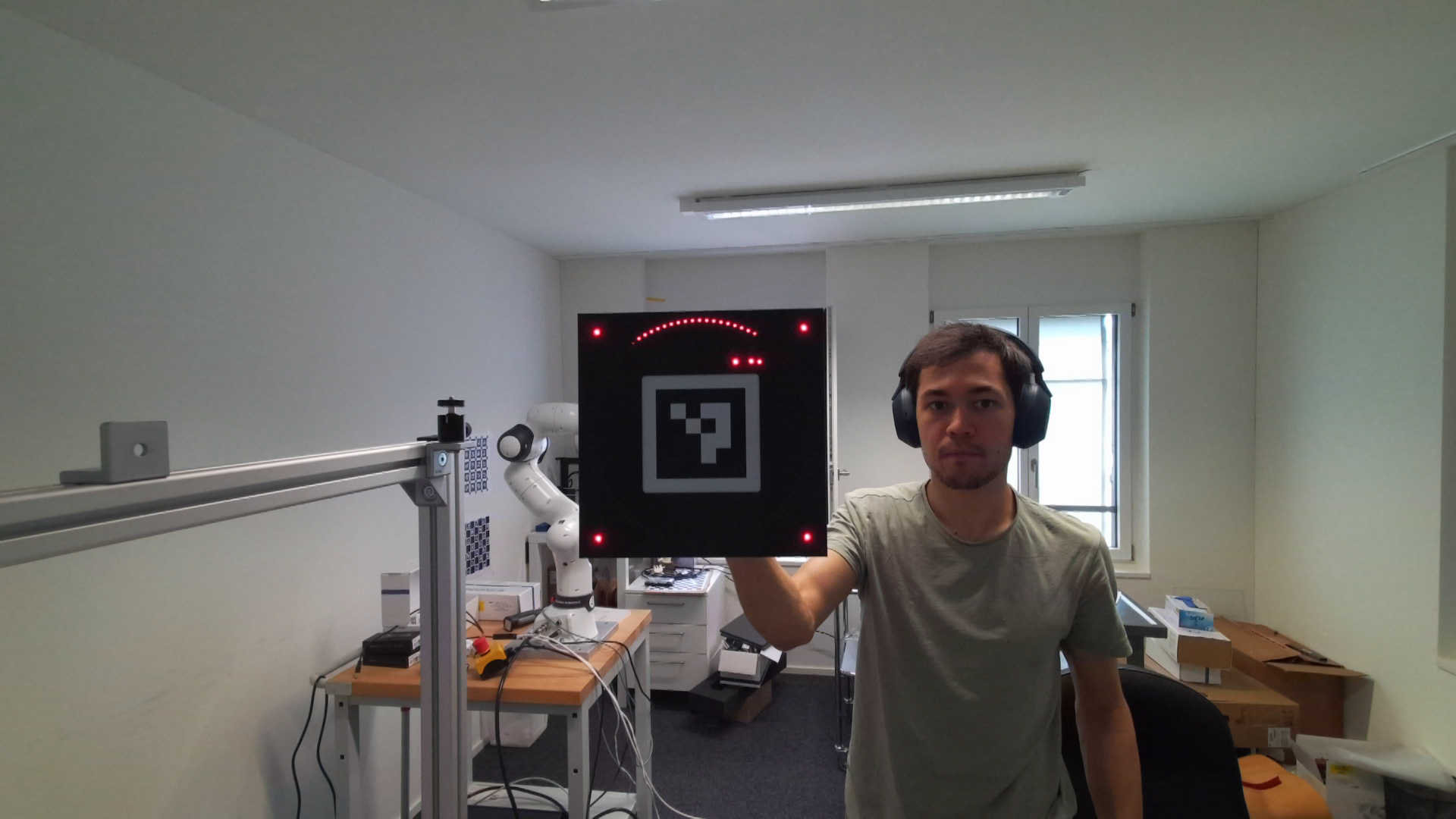}\label{fig:rocsync_in_use}}
\hspace{0.5cm}
\subfloat[\centering]{\includegraphics[height=3.3cm]{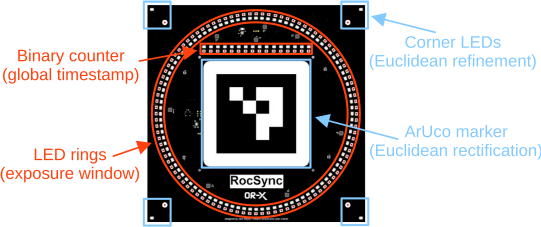}\label{fig:rocsync_labeled}}
\caption{Our custom \textit{LED Clock}. (\textbf{a}) Example of the device in use. (\textbf{b}) Rendering of the final PCB design with annotated features (red features encode the timestamp and blue features are used for detection \& Euclidean rectification).
The exposure window of the camera is visible in (a) as a red elliptical arc, which can be decoded with an accuracy of about 1 ms to enable sub-frame temporal synchronization. \label{fig:rocsync_overview}}
\end{figure}

% where our method fits
To address this gap, we propose a low-cost, portable, and camera-agnostic method for achieving millisecond-level temporal alignment across heterogeneous camera systems, ranging from inexpensive action cameras to high-end optical tracking systems. The proposed solution is designed to be scalable, supporting configurations ranging from as few as two cameras to several dozen, and remains robust in real-world scenarios.

Central to our approach is a custom \textit{LED Clock}, a $25\times25\,\text{cm}$ device that visually encodes timestamps using LEDs (see Figure~\ref{fig:rocsync_in_use}). An operator sequentially positions the LED Clock in front of each camera. Timestamps are encoded via two complementary LED components (Figure~\ref{fig:rocsync_labeled}): a binary counter providing a global timestamp with 100~ms resolution, and a circular arrangement of LEDs illuminated sequentially at 1~ms intervals. Because camera sensors integrate light over a non-zero exposure time, the circular sequence appears in the image as an elliptical arc, whose start and end points indicate the beginning and end of the exposure window. For accurate spatial decoding, an ArUco marker~\cite{GARRIDOJURADO20142280} of known geometry is included at the center, enabling detection and homographic rectification of the captured LED components (i.e., binary counter and elliptical arc). Combining the binary timestamp with the arc-based exposure estimate yields a millisecond-accurate global timestamp for the exposure window. This process is repeated for at least two frames per video, enabling estimation of both offset and drift for each camera and allowing precise temporal alignment of the entire recordings. In practice, the LED Clock is shown at the beginning and end of a recording to minimize extrapolation errors.

This approach allows multiple video streams to be synchronized with millisecond-level accuracy, regardless of the cameras’ native synchronization capabilities and scene content. Combined with established frame interpolation techniques~\cite{kye2025acevficomprehensivesurveyadvances, jin2025unifiedarbitrarytimevideoframe}, it opens up the possibility of synthesizing temporally aligned frames across all viewpoints, facilitating high-quality 3D reconstruction of dynamic scenes using unsynchronized consumer-grade cameras.

\paragraph{Contributions}
Our contributions are as follows: (1) We present a novel method for synchronizing heterogeneous camera systems comprising a combination of different infrared and RGB(-D) cameras. (2) We thoroughly evaluate our approach through a series of experiments conducted in both controlled laboratory settings and in real-world scenarios involving up to 25 cameras. (3) We demonstrate the importance and benefits of sub-frame temporal synchronization. (4) We contribute to the scientific community by open-sourcing all software and hardware designs developed in this work.

\section{Related Work}\label{sec:related_work}
Video synchronization methods can be broadly categorized into three groups: hardware-based synchronization, timecode-based synchronization, and content-based synchronization.

% -------------------------------------------------------------------------

\subsection{Hardware-based Synchronization}
Hardware-based synchronization uses an external electrical signal to trigger the camera shutters simultaneously across multiple devices. This ensures that all cameras capture frames at precisely the same moment. Typically, one camera or a dedicated device acts as the master and generates the synchronization signal, which is then distributed to all other cameras over dedicated cables. The signal distribution can follow either a star or daisy-chain topology. An example of a device that supports hardware synchronization is the Azure Kinect DK \cite{azure_kinect_multi_sync} which is widely used in scientific applications \cite{Hein2024CVPR, HEIN2025103613, ozsoy2025egoexoregoexocentricoperatingroom, ozsoy2025mmorlargemultimodaloperating}.

There have been approaches to replace dedicated synchronization cables with software-based triggering mechanisms guided by network time synchronization protocols over standard Ethernet \cite{s18041276, 10.1117/12.589840}. Similarly, wireless-aware time synchronization protocols over Wi-Fi have been proposed, eliminating the need for physical cabling while still achieving microsecond-level accuracy \cite{10109127}.

While hardware-based synchronization can achieve perfectly synchronized frames with microsecond-level accuracy, it is inherently limited to devices that support such methods, and interoperability between different camera systems is often not possible. Moreover, most setups require wired connections, which makes them unsuitable for mobile or wearable applications (e.g., head-mounted or vehicle-mounted cameras) and complicates installation, especially when deploying large-scale systems with dozens of cameras.

% -------------------------------------------------------------------------

\subsection{Timecode-based Synchronization}
In timecode-based synchronization, each camera independently records timestamp metadata either at the start of the capture or on a per-frame basis. These timestamps can subsequently be used to establish a temporal correspondence across multiple recordings. The timecodes may be derived from an internal real-time clock that has been synchronized prior to capture, or supplied by an external timecode generator, typically via an audio channel.

Common methods for synchronizing the internal real-time clock include network time protocols (e.g., NTP, PTP), GPS time, and dynamic QR codes (as implemented by GoPro~\cite{GoPro_HERO12_TimecodeSync}). The official GoPro documentation states a synchronization accuracy of $<50\,\text{ms}$ when using their dynamic QR codes~\cite{GoPro_HERO12_TimecodeSync}. Each camera then independently timestamps recorded videos using its internal clock. However, the shutters themselves are not synchronized across cameras.  As a result, residual misalignment of up to half a frame can remain even when pairing the closest timecodes between two video streams. Furthermore, since timecodes are typically stored with frame-level precision, sub-frame exposure timing is unknown, preventing interpolation between consecutive frames to achieve improved alignment.

Although these methods are easier to deploy than hardware synchronization and some initial synchronizations such as PTP or GPS time require no additional user intervention, they are generally much less precise due to clock drift over time. GoPro~\cite{GoPro_HERO12_TimecodeSync} reports that their cameras develop a noticeable time offset as a result of drift 60–90~minutes after synchronization. Timecode-based synchronization is widely used in professional video production, where small offsets of a few frames are often acceptable. However, scientific multi-view applications often require higher temporal accuracy and such inaccuracies are generally unacceptable. Nevertheless, it is frequently employed in dataset acquisition, especially in setups involving action cameras~\cite{li2022neural3dvideosynthesis}.

% -------------------------------------------------------------------------

\subsection{Content-based synchronization}
Content-based synchronization methods operate as post-processing techniques that analyze the actual image and/or audio content to establish temporal alignment between streams. This approach offers greater flexibility, as it does not require specialized cameras or specific setup steps prior to recording, since synchronization is performed offline after the recordings are complete. Synchronization cues can arise from naturally occurring content in the scene, or from deliberately introduced markers such as flashes, claps, or visual patterns. Similar to timecode-based synchronization, these methods do not synchronize the actual camera shutters. However, some content-based approaches provide timestamps with sub-frame precision~\cite{10.1007/978-3-642-32717-9_27, Liu_Ai_Xing_Li_Wang_Tao_2024}, enabling interpolation between frames to achieve improved alignment. Content-based methods can be further divided into several subcategories:

\paragraph{Audio-based synchronization}
One of the earliest and most straightforward methods, audio-based synchronization, leverages the audio track captured by cameras with a built-in microphone. Temporal alignment is then established by cross-correlating these signals or detecting and matching distinct events and peaks~\cite{10.1145/1291233.1291367}. Such cues are often intentionally introduced at the beginning of a recording by clapping or producing other sharp sounds. This allows video streams to be synchronized entirely independently of the visual content, making audio-based synchronization particularly robust in visually challenging environments such as cluttered scenes with repetitive textures, strong motion blur, or low-light conditions.

Although overlapping fields of view are not required, cameras must still be in close proximity to capture the same audio signals. Moreover, due to the relatively low propagation speed of sound, each camera should be positioned at roughly equal distances from the source; even a 5-meter difference introduces a 15\,ms discrepancy in time of arrival. As a result, this approach severely restricts the camera geometry and is not suitable for particularly noisy environments or outdoor settings where audio signals are weak or heavily distorted. Moreover, since audio and video are recorded, processed, and encoded independently, there can be an inherent offset between the two. As a result, even if the audio tracks of two cameras are perfectly synchronized, their corresponding video streams may still exhibit temporal misalignment. While audio is typically recorded at much higher sampling rates than video, enabling timestamping with sub-frame precision, this does not necessarily translate into accurate video alignment due to the aforementioned limitations. Despite all its limitations, its ease of use has led to its adoption in many applications \cite{ozsoy2025mmorlargemultimodaloperating, ozsoy2025egoexoregoexocentricoperatingroom, NEURIPS2024_c1017d0a}.

\paragraph{Visual feature correspondences}
This approach uses moving image features, such as key points, edges, or textures, to find correspondences between multiple video streams that share the same visual content \cite{6375020, shin2025audioposegeneralpurposeframework, 10.1007/978-3-642-32717-9_27}. While this makes this category of methods straightforward to use, performance highly depends on the content of the scene and the method requires overlapping camera views. There have also been approaches to combine audio and visual features for increased robustness \cite{5332301}.

Visual features can also include abrupt changes in lighting, such as a camera flash or the switching on or off of room lights, which create distinctive changes observable across multiple camera views. Furthermore, sub-frame synchronization accuracy can be achieved by exploiting rolling shutter artifacts present in many CMOS cameras \cite{smid2017rolling}. This approach relies on the assumption that the lighting change is captured simultaneously by all cameras, thus requiring them to be positioned within the same indoor environment. However, in our experiments, we observed that many lights do not switch on instantaneously, and in large environments such as operating rooms, the ceiling illumination often consists of multiple physical light sources that do not activate at exactly the same time, making it difficult to detect a distinct peak.

\paragraph{Pose matching}
In videos featuring human subjects, synchronization can be achieved by analyzing and matching human pose and movements across multiple camera views \cite{9022391, Liu_Ai_Xing_Li_Wang_Tao_2024}. This approach is fully automated and does not require any additional synchronization steps for recordings that already include moving humans. However, it requires overlapping fields of view and comparable zoom levels across cameras to ensure that the subject remains fully visible in all views.

\paragraph{Joint geometric calibration and temporal synchronization}
External calibration and temporal synchronization can be performed in one step by recording a moving checkerboard and jointly optimizing the camera extrinsics along with the temporal offset \cite{Albl_2017_CVPR, Hein2024CVPR}. While this approach can achieve accurate synchronization under ideal conditions, it suffers from several limitations: (i) it is operator-dependent, since the accuracy of temporal synchronization is sensitive to the checkerboard motion, and ambiguous checkerboard trajectories can lead to degenerate solutions in the joint optimization, (ii) it is computationally expensive, (iii) it relies on overlapping fields of view, and (iv) it degrades in performance when cameras have substantially different zoom levels (pattern too small or too large in one camera) \cite{fluckiger2025}.

\paragraph{Visual timestamp encoding}
Some approaches introduce dedicated hardware or visual signals to encode global timestamps directly into video frames. This can be achieved using dynamic QR codes, moving objects, blinking patterns, or LED arrays, which are presented to all cameras and decoded during post-processing \cite{s24216975, grauman2024egoexo4dunderstandingskilledhuman}. The decoded timestamps serve as precise anchor points for synchronizing camera streams to a global timeline. These methods have the advantage of not relying on overlapping camera views and functioning independently of scene content. However, this category of methods remains relatively unexplored, lacks standardization, and is not yet widely adopted.

% -------------------------------------------------------------------------

\subsection{Summary}
While both hardware-based and content-based synchronization techniques are widely used, each comes with inherent limitations. Hardware-based synchronization requires specialized equipment and complex setups, while content-based methods generally offer lower accuracy or only work for specific use-cases and completely lack support for infrared cameras. Additionally, content-based methods often have camera setup restrictions, as they require overlapping fields of view.

% manual, mention scale
Our method aims to bridge this gap by introducing a general-purpose LED Clock that uses visible and infrared LEDs to encode global timestamps with millisecond-level accuracy. Unlike prior solutions, it supports all RGB and infrared cameras, including 3D optical tracking systems, without requiring shared fields of view or special hardware integration. In addition, it provides millisecond-level synchronization to enable sub-frame interpolation.

\section{Method}\label{sec:method}
In this section, we first formalize the video synchronization problem. We then describe the key principles underlying our proposed approach.

% -------------------------------------------------------------------------

\subsection{Problem Formulation}\label{subsec:problem}
Multi-view video synchronization can be formulated as the problem of computing global timestamps for every frame in each video stream. We model the temporal behavior of a camera as a linear relationship between the local clock and some global reference clock, characterized by an initial offset and a constant drift factor. The initial offset arises because the recording starts at an arbitrary time offset relative to the global reference clock. In addition, each stream exhibits a constant drift component arising from small frequency deviations in the cameras’ internal oscillators. These variations cause the clocks to run at slightly different rates, which under the assumption of constant drift results in a temporal error that accumulates linearly over time. Any nonlinear effects, such as those induced by temperature fluctuations, are considered negligible for our target application. We therefore model the transformation from local to global time as follows:

\begin{linenomath}
\begin{equation}
T_i^{j} = \alpha_j \cdot t_i^{j} + \beta_j,
\label{eq:linear_model}
\end{equation}
\end{linenomath}
where $T_i^{j}$ is the global timestamp of the $i$-th frame from camera $j$, $t_i^{j}$ is the local timestamp, $\alpha_j$ represents the clock drift of camera $j$, and $\beta_j$ denotes its initial time offset. We formulate temporal synchronization of camera $j$ as the problem of estimating $\alpha_j$ and $\beta_j$ from two or more measured samples $(t_i^{j}, \hat{T}_i^{j}) \in \mathcal{S}_j$, where each sample represents a measurement of the global clock $\hat{T}_i^{j}$ at a specific local timestamp $t_i^{j}$. We formulate this as a linear least-squares problem that minimizes the error between the predicted and observed global timestamps:
\begin{linenomath}
\begin{equation}
(\alpha_j, \beta_j) = \arg\min_{\alpha_j, \beta_j} \sum_{(t_i^{j}, \hat{T}_i^{j}) \in \mathcal{S}_j} \left( \hat{T}_i^{j} - (\alpha_j \cdot t_i^{j} + \beta_j) \right)^2,
\label{eq:argmin_camera}
\end{equation}
\end{linenomath}
which can be efficiently solved using the well-known closed-form solution~\cite{doi:10.1137/1.9781421407944} or any more advanced robust linear regression method~\cite{10.1214/aoms/1177703732}. If only a single sample is available, the system can still be solved by setting $\alpha_j = 1$, which assumes that the camera's local clock has no drift relative to the global clock.

% -------------------------------------------------------------------------

\subsection{Proposed Method}
Our method introduces a custom LED Clock equipped with an array of LEDs that serves as a global reference clock (see Figure~\ref{fig:rocsync_overview}). This device continuously updates its LED pattern to visually encode the current timestamp with millisecond precision. These timestamps can then be extracted from video frames in which the LED Clock device is visible.

Our LED pattern is divided into two subsets (see Figure~\ref{fig:rocsync_labeled}). The first subset consists of 100 LEDs arranged in a circle, with exactly one LED illuminated at a time. This active LED advances one position every millisecond, completing a full rotation in 100~ms. Because camera shutters are not instantaneous, light is integrated over the exposure period (typically a few milliseconds), causing several adjacent LEDs to appear illuminated in a single frame. The first and last LED in this arc correspond to the start and end of the exposure window, respectively, enabling precise determination of both timestamps (see Figure~\ref{fig:pipeline3}). The step duration of 1~ms was chosen to provide millisecond-level accuracy, while the use of 100 LEDs ensures correct operation for exposure times of up to 100~ms. Together, these design choices ensure compatibility with a wide range of cameras, supporting frame rates from 10 to 1000~fps. Since the LED ring wraps around every 100~ms, a second subset of LEDs is arranged in a linear row to serve as a binary counter. This counter increments with each full rotation of the ring, keeping track of the number of complete 100~ms revolutions. Together, the ring and binary counter LEDs uniquely encode the current timestamp with millisecond precision.

Each frame that captures the clock device can be decoded into a pair of timestamps $(\hat{T}^j_{i,start}, \hat{T}^j_{i,end})$, representing the start and end of the exposure of the $i$-th frame from camera $j$. We assume that the start of the exposure is always aligned with the regular frame interval, while automatic exposure control only affects the end. Consequently, $\hat{T}^j_{i,start}$ is used as the measured timestamp $\hat{T}_i^{j}$ when solving Equation~\ref{eq:argmin_camera}. By briefly showing the clock device to each camera, we obtain a set of samples
\[
\mathcal{S}_j = \{ (t_{i_1}^j, \hat{T}^j_{i_1,start}), ..., (t_{i_N}^j, \hat{T}^j_{i_N,start}) \},
\]  
which are used to estimate the drift and offset parameters of camera $j$, as described in Section~\ref{subsec:problem}.
In practice, the LED Clock is recorded at both the beginning and the end of each capture session, establishing two temporal reference points that minimize estimation error and account for potential clock drift over long recordings.

\section{Implementation}\label{sec:impl}
Our method consists of two main components: the LED Clock device and a processing pipeline for decoding the global timestamps from video frames and estimating the offset and drift for each video stream.

% -------------------------------------------------------------------------

\subsection{Hardware}
The custom clock device is based on a $25\times25\,\text{cm}$ PCB featuring 100 red and IR LEDs arranged in a circle, complemented by 16 additional red and IR LEDs serving as a binary counter (see Figure~\ref{fig:rocsync_labeled}). This design enables the encoding of unique timestamps with millisecond precision for up to 109~minutes, a limit that can be easily extended by adding more LEDs to the binary counter, depending on the application requirements.

The PCB dimensions were chosen to ensure sufficient spacing between LEDs, minimizing light interference from neighboring LEDs, which could otherwise introduce decoding errors, particularly when the device is viewed from a far distance or captured in low-resolution video. At the same time, the size is small enough to be fully captured by cameras with very narrow fields of view, such as zoom cameras focused on a specific point of interest. The resulting $25\times25\,\text{cm}$ size ensures that the clock is compatible with a wide range of camera fields of view and operator distances.

It features a central ArUco marker~\cite{GARRIDOJURADO20142280} for robust detection in cluttered environments and initial homography rectification of the imaged PCB. The marker is complemented by four corner LEDs that serve as precise, well-spaced point correspondences to refine the homography rectification provided by the corners of the ArUco marker. One corner includes an additional IR LED to resolve rotation ambiguity in infrared videos where the ArUco marker is not visible.

The entire system is controlled by a 32-bit RISC-V microcontroller with an external temperature-compensated crystal oscillator (TCXO) for precise timekeeping and is powered by a rechargeable battery, providing several hours of independent operation per charge.

The PCB was designed using the open-source EDA software \textit{KiCAD}\footnote{\url{https://kicad.org}}, while auxiliary mechanical components were modeled in \textit{Fusion360} and subsequently 3D printed. To maximize contrast between illuminated and non-illuminated LEDs under diverse lighting conditions, the PCB was wrapped in a thin layer of matte black film, effectively reducing reflections from strong light sources. Component selection was optimized for manufacturing by the PCB fabrication service JLCPCB\footnote{\url{https://jlcpcb.com}}.

The hardware design prioritizes cost-effectiveness and accessibility. The final device can be manufactured and assembled using standard PCB fabrication services accessible to individual consumers, with an estimated cost of approximately US\$20 per unit.

% -------------------------------------------------------------------------

\begin{figure}[H]
    \centering
    \subfloat[\centering]{\includegraphics[height=3.5cm]{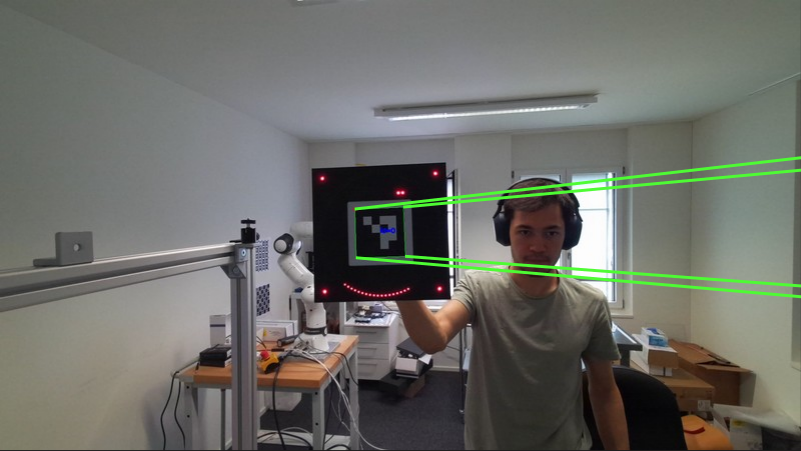}\label{fig:pipeline1}}
    \subfloat[\centering]{\includegraphics[height=3.5cm]{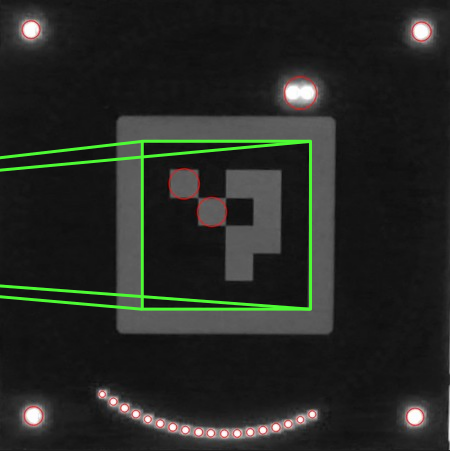}\label{fig:pipeline2}}
    \subfloat[\centering]{\includegraphics[height=3.5cm]{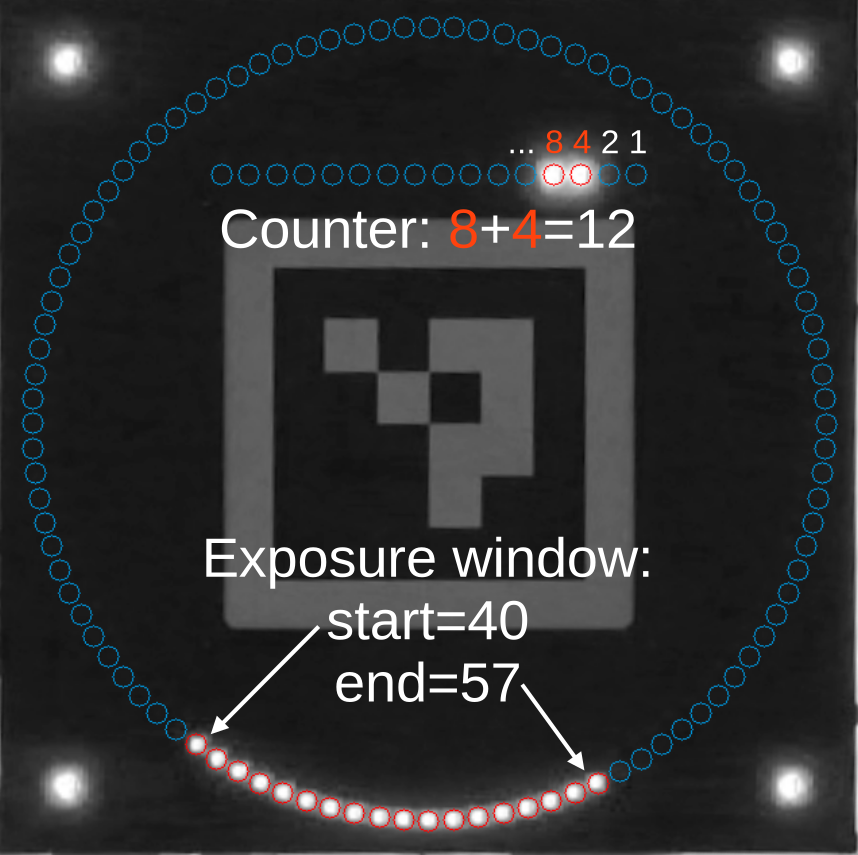}\label{fig:pipeline3}}
    \caption{Our computer vision pipeline: (\textbf{a}) raw image with ArUco detection, (\textbf{b}) coarse homography reprojection using the ArUco marker with detected blobs shown in red circles, and (\textbf{c}) decoded LEDs after refined homography reprojection using the corner blobs from the previous step. All LED positions are marked with blue circles, while illuminated LEDs are highlighted in red. In this example, the binary counter shows 12 and the illuminated ring segment indicates that the exposure was taken from LED 40 to LED 57. Combining this information, we know that the exact exposure of this frame occurred from $12*100+40=1240\,\text{ms}$ to $12*100+57=1257\,\text{ms}$.\label{fig:rocsync_usage}}
\end{figure}

\subsection{Software}
\label{subsec:software}
The software component consists of a Python module that reads a video, analyzes each frame, and outputs the estimated offset and drift values for the entire video. Using these measurements, global timestamps can then be recomputed for every frame using Equation~\ref{eq:linear_model}. The following steps describe our computer vision pipeline (see Figure~\ref{fig:rocsync_usage} for a visualization of the main steps, from left to right):
\begin{enumerate}
    \item \textbf{Find ArUco marker:} Detect the central ArUco marker to determine the approximate position and orientation.
    \item \textbf{Coarse reprojection:} Use the detected marker's corners to perform a coarse planar homography rectification of the imaged clock.
    \item \textbf{Locate corner LEDs:} Identify the corner LEDs in the coarsely rectified image.
    \item \textbf{Accurate reprojection:} Use the extracted corner LEDs to refine the homography rectification for improved accuracy. While ArUco marker detection is generally robust, the detected corner positions are not always pixel-perfect, making this refinement necessary.
    \item \textbf{Decode LEDs:} Decode the circle and binary counter LEDs by thresholding their respective areas to obtain exact timestamps for the start and end of the exposure.
    \item \textbf{Timestamp fitting:} Perform robust linear regression on all extracted timestamps to reject outliers and estimate global timestamps for all frames (solving Equation~\ref{eq:argmin_camera}).
\end{enumerate}

% \paragraph{Robustness}
To ensure robustness, the pipeline incorporates specific rejection criteria at multiple levels:

In \textit{step 1}, a frame is rejected if the detected ArUco marker occupies less than 0.2\% of the image area. This condition indicates that the LED Clock was held too far from the camera, resulting in insufficient resolution to reliably distinguish adjacent LEDs and therefore leading to unreliable decoding.

In \textit{step 3}, a frame is rejected if the corner LEDs in the coarse reprojection deviate from their expected positions by more than a specified threshold.

In \textit{Step 5}, dynamic local thresholding is used to handle reflections and uneven illumination. Additionally, a frame is rejected if the ring does not display a single illuminated sector or if the sector crosses the counter boundary, as this indicates the counter changed during the exposure, resulting in an ambiguous reading which cannot be decoded correctly.

In \textit{step 6}, robust RANSAC regression~\cite{10.1145/358669.358692} is applied to combine the independent timestamps from all decoded frames by fitting a linear function and computing the global offset and drift of the video. This approach effectively discards outlier frames where the clock was decoded incorrectly, producing timestamps that do not conform to the assumed linear model. RANSAC was chosen over other robust linear regression methods because it explicitly excludes outliers when estimating model parameters, rather than merely down-weighting them.

Together, these precautions ensure reliable performance even under suboptimal conditions, such as poor lighting, motion, or the board being viewed from a distance or at an angle.

\section{Experiments}
To evaluate the accuracy and robustness of our method, we designed a comprehensive testing framework that compares it against existing synchronization techniques across various camera configurations and combinations, both in laboratory and real-world scenarios.
For each configuration, we designed specific experiments and selected an appropriate quantitative metric based on the capabilities and limitations of the cameras involved.

In our first experiment, we used two Kinect cameras with hardware synchronization as the ground truth reference. This setup allows us to compute the exact residual synchronization error in the time domain.

In our second experiment, we test the wide applicability of our method by comparing it with existing synchronization methods for consumer-grade cameras. Since GoPro cameras lack support for hardware synchronization to provide a reliable ground-truth reference, we instead performed an external calibration for every synchronization method using a fast-moving checkerboard. The resulting reprojection error was then used as a practical metric for evaluating temporal alignment, as it is minimized when the video streams are perfectly synchronized, thus providing a direct and quantitative measure of the temporal alignment achieved by each approach. Additionally, we used the sub-frame accurate timestamps provided by our method to interpolate synchronized 2D positions of the checkerboard instead of simply using the closest frame for every video stream, further minimizing the reprojection error.

Then, in our third experiment, we evaluated the accuracy of our method across two different modalities, namely RGB video cameras and a professional-grade IR optical tracking system. This setup enabled us to assess the robustness and generalizability of our approach when applied to heterogeneous camera systems.

In the fourth and fifth experiments, we demonstrated the importance of sub-frame accurate synchronization for multi-view computer vision tasks. Specifically, we synchronized a multi-view recording with our method and compared the results of 3D pose estimation and 3D reconstruction when using sub-frame interpolation versus nearest-frame matching.

Finally, in the sixth experiment, we tested robustness outside controlled laboratory conditions through a case study in large-scale surgical data collection involving 25 cameras.
The data collection featured a highly heterogeneous camera array, comprising devices from various manufacturers and price ranges, with differing settings such as field of view, frame rate, shutter speed, and sensor type, demonstrating the broad applicability of our method.

Together, this rigorous evaluation allowed us to thoroughly assess the accuracy and versatility of our method across diverse laboratory and real-world scenarios.

% -------------------------------------------------------------------------

\subsection{Experiment 1: Validation against hardware synchronization}
In this experiment, we evaluate the absolute synchronization error of our method by comparing its output against a ground-truth reference. To obtain the ground truth, we leveraged the hardware synchronization capabilities of the Azure Kinect DK (Microsoft Corporation, Redmond, WA, USA), assuming it provides perfect synchronization. The recording procedure was designed to closely match real-world multi-view data collections. We applied our method to recompute the timestamps for both video streams separately and then calculated the root mean square error (RMSE) between them across all frames. This error should ideally be zero because the shutters are perfectly synchronized. However, it is important to note that the Microsoft Azure Kinect DK uses an \texttt{OV12A10} RGB sensor with a rolling shutter, meaning that different image rows are exposed sequentially in time. Consequently, the LED pattern on the clock may differ slightly between frames (see the Results paragraph for details). Together with the relatively low frame rate of 30 fps, this setup constitutes a challenging scenario for hardware-synchronization-free methods in terms of synchronization accuracy.

\paragraph{Experimental setup}
Two Kinect cameras were placed next to each other, facing opposite directions, and connected using a cable to synchronize their shutter. We then used \verb|k4arecorder| on a central laptop to record from both cameras simultaneously. The cameras were set to a resolution of $1920 \times 1080 \,\text{pixels}$ at 30~fps with a fixed exposure time of 8.33~ms.

The recording followed a protocol representative of real-world, non-laboratory conditions, similar to those in large-scale data collections involving many cameras (e.g., the 25-camera setup used in Experiment 6), where not all cameras share overlapping visual content.
We first showed the clock to the first camera for approximately 5 seconds, then waited 2 minutes, and subsequently showed it to the second camera for 5 seconds. The 2-minute delay simulates the scenario of synchronizing more than two cameras, assuming that one full cycle of showing the clock to every camera takes approximately 2~minutes. The clock was positioned approximately 150~cm from the camera and centered in the field of view. After the initial synchronization cycle, the actual procedure of interest was simulated by waiting for 5~minutes before performing a second identical synchronization cycle. This allowed for a more accurate measurement of the drift over time. This process was repeated five times, resulting in five independent 14-minute recordings from two cameras.

\paragraph{Evaluation score}
The 10 videos were processed independently using our method to recompute global timestamps for every frame. Finally, for every video pair, we computed the root mean square error (RMSE) between the recomputed timestamps. The RMSE is defined as:
\[
\text{RMSE} = \sqrt{\frac{1}{N} \sum_{i=1}^{N} \left( t_i^{1} - t_i^{2} \right)^2}
\]
where \(N\) is the total number of frames, and \(t_i^{1}\) and \(t_i^{2}\) represent the recomputed timestamps of the \(i\)-th frame in the first and second video streams, respectively.

\begin{table}[H]
\centering
\caption{Root mean square error (RMSE) in milliseconds between the two video streams.}
\begin{tabular}{ccccc|c}
\toprule
\textbf{Run 1} & \textbf{Run 2} & \textbf{Run 3} & \textbf{Run 4} & \textbf{Run 5} & \textbf{Mean} \\
\midrule
0.38 & 1.30 & 1.94 & 1.73 & 1.37 & \textbf{1.34} \\
\bottomrule
\end{tabular}\label{table:rmse_gt}
\end{table}

\paragraph{Results} RMSE values over the five runs are reported in Table \ref{table:rmse_gt}, with an average RMSE of 1.34~ms, indicating strong agreement with the ground truth reference. This error is significantly lower than the duration of a single frame at 30~fps (33.3~ms), demonstrating that our method achieves meaningful sub-frame accuracy. We attribute the residual error primarily to the rolling shutter of the Kinect's RGB sensor. Unlike a global shutter that exposes all pixels simultaneously, the rolling shutter captures each row sequentially over a short interval. This causes slight temporal offsets across the frame, meaning that the effective exposure time for a given pixel depends on its vertical position. Therefore, the timestamp encoded by the clock appears slightly shifted depending on the clock’s position within the frame. Additionally, since the LEDs are arranged in a ring and therefore at different vertical positions, the decoded timestamp can vary slightly based on which specific LED was illuminated during exposure, for example, whether the observed arc is predominantly horizontal or vertical. This is further supported by the slight variations in exposure times shown in Figure~\ref{fig:kinect_rollingshutter}. To reduce this artifact, the clock can be held at a greater distance from the camera, effectively decreasing the vertical pixel distance between the highest and lowest LED.

% -------------------------------------------------------------------------

\subsection{Experiment 2: GoPro to GoPro}
In this experiment, we compare our method to existing synchronization approaches for aligning two GoPro video streams. Since GoPro cameras do not offer built-in hardware synchronization that can serve as a ground truth reference, we evaluate performance indirectly by performing stereo external calibration based on a moving checkerboard and reporting the resulting reprojection error. The underlying intuition is that when video streams are accurately synchronized, the reprojection error is minimized, thus providing a practical and quantitative measure of temporal alignment. We deliberately maintained the checkerboard in continuous rapid circular motion to amplify the reprojection error caused by temporal misalignment, making the results easier to interpret. We are comparing the following methods:
\begin{enumerate}
    \item RocSync (our method)
    \item Global illumination-based synchronization (toggling room lights)
    \item Audio-based synchronization (clapping) using \textit{AudioAlign}~\cite{6424693}
    \item Timecode-based synchronization (GoPro Timecode Sync using QR code)
\end{enumerate}

For global illumination-based synchronization, we implemented a simple script that estimates the temporal offset by cross-correlating the gradient of average global illumination between the two cameras. Audio-based synchronization was performed with the open-source software \textit{AudioAlign}~\cite{6424693} by applying FP-EP followed by cross-correlation. Like most battery-powered cameras, GoPros include an internal real-time clock that maintains wall time. This clock is then used to generate per-frame timestamps with frame-level precision, which are encoded in the \textit{SMPTE timecode} format. These timestamps can be used to identify the corresponding frames in multiple video streams. The \textit{GoPro Labs Firmware}~\footnote{\url{https://gopro.com/en/us/info/gopro-labs}} enables fast and accurate timecode synchronization using QR codes displayed on a smartphone.

\paragraph{Experimental setup} We placed two GoPro cameras (GoPro Hero 12 and GoPro Hero 10) approximately 1~meter apart to capture the same scene. Both cameras were running the \textit{GoPro Labs Firmware} and they were configured to record at a resolution of $3840\times2160$ with 60~fps and a fixed exposure time of 1/480~s. The internal real-time clocks of the cameras were synchronized immediately prior to recording using an animated QR code displayed on a phone, a feature of the \textit{Labs Firmware}.

The recording was then started manually on both cameras, after which we performed the calibration procedures required for each synchronization method. Specifically, we clapped, turned the room light off and on, and finally showed our LED Clock to each camera for 5~seconds. After completing these initial calibration steps, a $270\times180\,\text{mm}$ 3D-printed precision checkerboard was continuously moved along a circular trajectory for 30~seconds, making sure that it remained visible in both camera views. Lastly, the initial calibration steps were repeated once more.

The Hero 12 was selected as the reference camera, and the Hero 10 was aligned to it using the different synchronization methods. For each method, the videos were synchronized independently, and the same set of 100 consecutive frames was then used for stereo external calibration with the \verb|cv2.stereoCalibrate()| function from OpenCV. In addition, all original videos were downsampled to 30~fps (half the native frame rate) using \textit{ffmpeg}, and the entire process was repeated on these recordings.

\paragraph{Evaluation score}
The synchronization accuracy of each method is evaluated using the mean reprojection error (MRE) of all checkerboard corners in the 100 consecutive frames after stereo calibration. For two cameras, a fully symmetric MRE can be defined using both forward and inverse projection. Let $\Pi_1$ and $\Pi_2$ denote the projection functions of the left camera and right camera respectively.
The MRE is written as:
\begin{equation}
\mathrm{MRE}
=
\frac{1}{2NM}
\sum_{i=1}^{N}
\sum_{j=1}^{M}
\Big(
\big\|
\mathbf{x}^{(1)}_{ij}
-
\Pi_1\!\left(\mathsf{R}_i,\ \mathbf{t}_i,\ \mathbf{X}_j\right)
\big\|_2
+
\big\|
\mathbf{x}^{(2)}_{ij}
-
\Pi_2\!\left(\mathsf{R}_{12}\mathsf{R}_i,\ \mathsf{R}_{12}\mathbf{t}_i+\mathbf{t}_{12},\ \mathbf{X}_j\right)
\big\|_2
\Big)
\end{equation}
where 
$N$ is the number of image pairs (views), 
$M$ is the number of checkerboard points, 
$\mathbf{x}^{(1)}_{ij}$ and $\mathbf{x}^{(2)}_{ij}$ are the observed image points in the left and right images, 
$\mathbf{X}_j$ is the 3D position of the $j$-th checkerboard point, 
$\mathsf{R}_i$ and $\mathbf{t}_i$ are the pose (rotation and translation) of the checkerboard in the left camera frame, 
and $\mathsf{R}_{12}$ and $\mathbf{t}_{12}$ are the rotation and translation from the left camera to the right one. This symmetric formulation measures the reprojection error in both directions, providing an assessment of calibration and temporal alignment.

\paragraph{Results} Table~\ref{tab:reprojection_error_gopro_combined} reports the mean reprojection error across five runs, comparing different synchronization methods at 30 fps and 60 fps. At 30~fps, our method with interpolation achieves the lowest errors overall, consistently outperforming existing approaches such as light- or audio-based synchronization, and far surpassing timecode alignment. Without interpolation, our method performs similarly to light-based synchronization, as both provide frame-accurate offset estimation. At 60~fps, the margin between our interpolated method and the closest competing approaches narrows, since the maximum error of frame-accurate synchronization is inherently bounded by half a frame interval. With interpolation, our method achieves nearly the same error at 30 fps as at 60 fps, clearly demonstrating its sub-frame accuracy.

It is important to note that this experiment represents an ideal scenario for light-based synchronization, with a single ceiling-mounted lamp that turns on and off nearly instantaneously. In many real-world settings and larger rooms, illumination often comes from multiple independent lamps and is neither synchronized nor uniform.

Sound-based synchronization exhibits a noticeable error, which we suspect arises from the fact that video and audio streams are encoded separately and only later combined in a container. Differences in audio/image pipeline latency can introduce timing offsets that vary across camera models. Because the error exceeds a single frame interval, sub-frame interpolation does not consistently improve the result.

\begin{table}[H]
\centering
\caption{Mean reprojection error (MRE, in pixels) of stereo external calibration across 5 runs. Values are shown as 30~fps / 60~fps. Lower values indicate better temporal alignment.}
\label{tab:reprojection_error_gopro_combined}
\small
\begin{tabular}{l|ccccc|c}
\toprule
\textbf{Method} & \textbf{Run 1} & \textbf{Run 2} & \textbf{Run 3} & \textbf{Run 4} & \textbf{Run 5} & \textbf{Mean} \\
\midrule
\textbf{RocSync (w/ interp.)}   & \textbf{0.91 / 0.80}& \textbf{1.02 / 0.87}& \textbf{0.83 / 0.71}& \textbf{1.06 / 1.11}& \textbf{1.28 / 1.26}& \textbf{1.02 / 0.95}\\
RocSync                         & 3.19 / 0.84& 3.31 / 1.08& 3.10 / 0.71& 3.93 / 1.10& 1.82 / 1.82& 3.07 / 1.11\\
Light      & 3.19 / 0.84& 3.31 / 1.08& 3.20 / 0.71 & 3.93 / 1.10& 1.82 / 1.82& 3.09 / 1.11\\
Audio (w/ interp.)              & 11.80 / 12.08& 13.17 / 18.18& 10.48 / 11.31& 14.82 / 15.08& 14.25 / 14.53& 12.90 / 14.64\\
Audio                           & 9.29 / 12.44& 10.28 / 13.54& 8.60 / 12.31& 11.97 / 16.17& 16.56 / 13.07& 11.34 / 13.91\\
Timecode                        & 50.85 / 49.14& 45.39 / 47.31& 60.21 / 57.86& 69.19 / 70.12& 47.36 / 51.26& 54.60 / 55.14\\
\bottomrule
\end{tabular}
\end{table}

% -------------------------------------------------------------------------

\subsection{Experiment 3: IR pose-tracking to RGB camera}
One of the advantages of our approach lies in its ability to synchronize heterogeneous camera systems, including those that rely on infrared imaging. To evaluate this capability, we tested the combination of a consumer action camera (GoPro Hero 12) and a high-end IR marker tracking system. The pose-tracking modality was represented by the Atracsys fusionTrack 500 (Atracsys LLC, Puidoux, Switzerland), a stereo infrared tracking system used for high-precision 3D localization of infrared-sensitive markers in surgical applications. To our knowledge, no existing hardware synchronization method supports this camera setup, so we use the reprojection error as a means to validate the proposed synchronization method.

\paragraph{Experimental setup} Both cameras were mounted on the same tripod. The GoPro camera was set to record at a resolution of $3840\times2160$ with 60~fps. Similar to the previous experiment, a $270 \times 180\,\text{mm}$ 3D-printed multimodal checkerboard with embedded infrared fiducials was moved through the scene with rapid linear motions visible to both cameras. Since the fusionTrack system records 3D marker and fiducial coordinates instead of 2D images, we developed a dedicated pipeline for detecting and decoding our LED Clock in these recordings (see Appendix~\ref{sec:ftk_pipeline}). After synchronizing the recordings using our method, we performed stereo extrinsic calibration with the multimodal checkerboard following the approach of Hein et al.~\cite{Hein2024CVPR}. The experiment was repeated three times in total.

\paragraph{Evaluation score} We use again the MRE. This time, it evaluates how accurately the corners of a moving 3D checkerboard rigidly attached to an IR marker, are reprojected onto their corresponding 2D detections in the RGB images. 
For each synchronized pair, the 3D points $\mathbf{X}_j$ correspond to the checkerboard corners expressed in the marker reference frame. 
Using the pose of the IR marker with respect to the IR camera $(\mathsf{R}_{IM,i}, \mathbf{t}_{IM,i})$, these points are first expressed in the IR camera frame, then transformed into the RGB camera frame using the calibrated extrinsic parameters $(\mathsf{R}_{12}, \mathbf{t}_{12})$, and finally projected into the RGB image through the pinhole model $\Pi_2(\cdot)$. 
The MRE is computed as follows:

\begin{equation}
\mathrm{MRE}
=
\frac{1}{N\,M}
\sum_{i=1}^{N}
\sum_{j=1}^{M}
\Big\|
\mathbf{x}_{ij}
-
\Pi_2\!\left(
\mathsf{R}_{12}
\big(\mathsf{R}_{IM,i}\mathbf{X}_j+\mathbf{t}_{IM,i}\big)
+
\mathbf{t}_{12}
\right)
\Big\|_2
\end{equation}
where 
$N$ is the number of synchronized IR--RGB pairs, 
$M$ is the number of checkerboard corners per view, 
$\mathbf{x}_{ij}$ are the 2D detections in the RGB camera (camera~2), 
$\mathbf{X}_j$ are the 3D coordinates of the checkerboard corners expressed in the marker reference frame, 
$\mathsf{R}_{IM,i}$ and $\mathbf{t}_{IM,i}$ represent the rotation and translation of the marker with respect to the IR camera (camera 1) for view~$i$, estimated from the IR stereo tracking system, 
$\mathsf{R}_{12}$ and $\mathbf{t}_{12}$ are the rotation and translation from the IR to the RGB camera, 
and $\Pi_2(\cdot)$ denotes the projection function of the RGB camera.

This formulation allows evaluating the overall alignment accuracy between both sensing modalities by directly comparing the projected positions of the 3D checkerboard corners to their detected 2D image locations in the RGB frames.

\paragraph{Results}
Table~\ref{tab:reprojection_error_heterogeneous} reports the mean reprojection error (MRE) across three runs. While no direct comparison to other synchronization methods is possible in this heterogeneous setup, the observed error of 1.81~px at 4K resolution is consistent with previously reported results for the same configuration, namely same checkerboard size and geometry, as well as comparable checkerboard-to-camera distances, using joint optimization of extrinsic parameters and time offset~\cite{HEIN2025103613}. This result demonstrates high synchronization accuracy.

\begin{table}[H]
\centering
\caption{Resulting mean reprojection error (MRE, in pixels) of stereo external calibration across 3 runs when synchronizing GoPro and Atracsys fusionTrack 500 using our method. Lower values indicate better temporal alignment.}
\label{tab:reprojection_error_heterogeneous}
\begin{tabular}{ccc|c}
\toprule
\textbf{Run 1} & \textbf{Run 2} & \textbf{Run 3} & \textbf{Mean $\pm$ Std} \\
\midrule
1.89 & 1.57 & 1.97 & \textbf{1.81 $\pm$ 0.17} \\
\bottomrule
\end{tabular}
\end{table}

% -------------------------------------------------------------------------

\subsection{Experiment 4: 3D pose estimation}
In this experiment, we evaluate the applicability of our method to a common real-world downstream task: multi-view 3D human hand pose estimation. We quantitatively assess performance by comparing triangulation error when using interpolation with sub-frame timestamps versus the nearest-frame strategy. Additionally, side-by-side visualizations of the triangulated geometries qualitatively demonstrate fewer abnormalities when using sub-frame interpolation.

\paragraph{Experimental setup}
We employed four GoPro Hero 12 cameras arranged in a square with approximately 30~cm spacing between adjacent units. All cameras recorded at 4K resolution and 30~fps with a shutter speed of 1/240. The captured sequence depicts a simulated surgical scenario in which a surgeon manipulates a saw bone.

Synchronization using our method was performed in two rounds following the same procedure as in previous experiments. For each view, 2D hand poses were manually annotated by an experienced annotator. Six distinct moments (denoted $T_1$ to $T_6$) with significant hand motion, where one hand was fully visible, were selected, and 3D hand joints were triangulated twice per moment: first by using the nearest frame from each view, and second by interpolating the 2D coordinates between the two nearest frames in each view using the sub-frame accurate timestamps obtained through our method.

In a second step, the four viewpoints were partitioned into two sets (GoPro 1 \& 2 and GoPro 3 \& 4), and moment $T_5$ was triangulated separately for each set using only two cameras (again with and without interpolation). The resulting 3D poses were then compared to assess the discrepancy between two independent 3D pose estimations, namely from the two camera subsets, corresponding to the same moment $T_5$.

\paragraph{Results}
For all six moments ($T_1$–$T_6$), we evaluated the triangulation error of the 3D hand joints in terms of the mean reprojection error. The results are reported in Table~\ref{tab:mre_summary2}. We obtain a mean triangulation error of 4.4~px when using interpolation, compared to 8.8~px when using the nearest-frame synchronization strategy. These results highlight the clear benefit of sub-frame accurate synchronization.

\begin{figure}[H]
\centering
\subfloat[\centering{Nearest frame}]{\includegraphics[width=5cm]{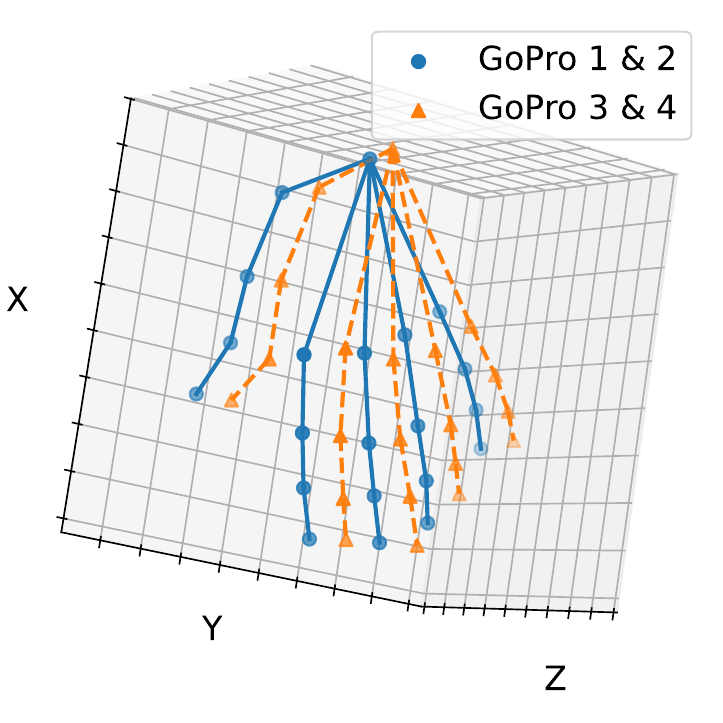}}
\hspace{0.5cm}
\subfloat[\centering{Sub-frame interpolation}]{\includegraphics[width=5cm]{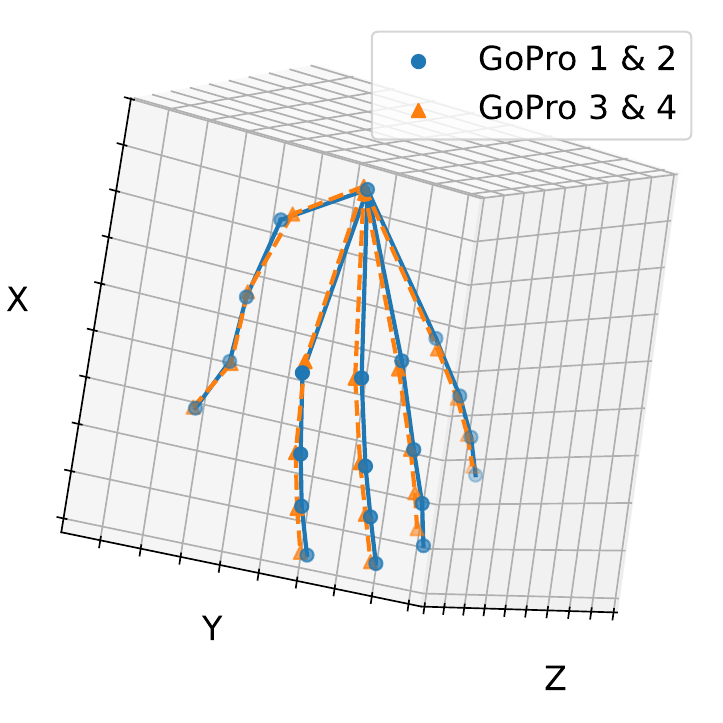}}
\caption{Mean Euclidean distance between two independent 3D hand pose reconstructions obtained from different camera subsets (GoPros 1\&2 and 3\&4) at $T_5$, (\textbf{a}) with and (\textbf{b}) without sub-frame synchronization. Sub-frame synchronization substantially reduces the discrepancy between reconstructions, highlighting its importance for accurate 3D hand pose estimation.\label{fig:euclidiane_distance}}
\end{figure}

In the partitioned sparse-view scenario, we evaluated the discrepancy between the two resulting 3D poses. As shown in Figure~\ref{fig:euclidiane_distance}, sub-frame synchronization reduced the mean Euclidean distance between poses from 26.07~mm without interpolation to 7.52~mm with interpolation, confirming its clear benefit.

\begin{figure}[H]
\centering
\subfloat[\centering{Nearest frame}]{\includegraphics[width=5cm]{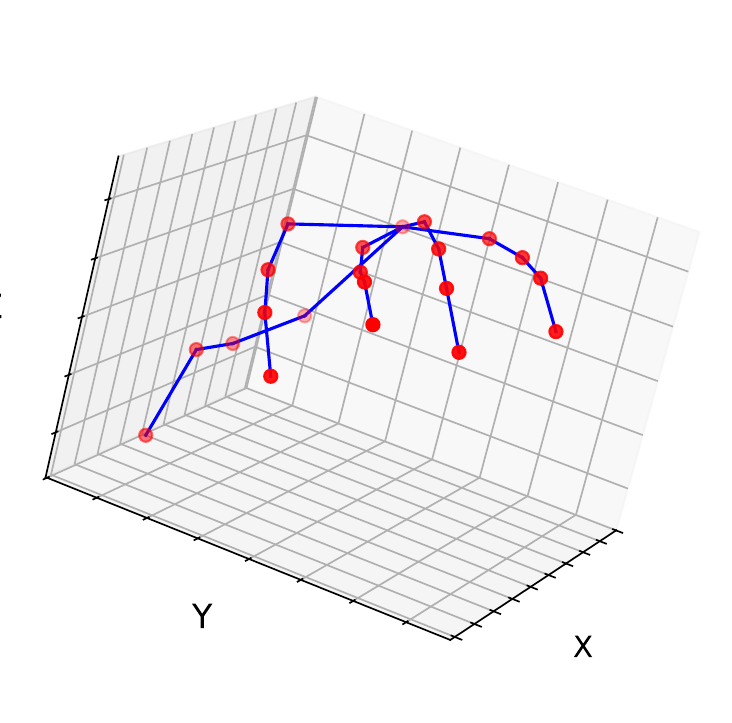}}
\hspace{0.5cm}
\subfloat[\centering{Sub-frame interpolation}]{\includegraphics[width=5cm]{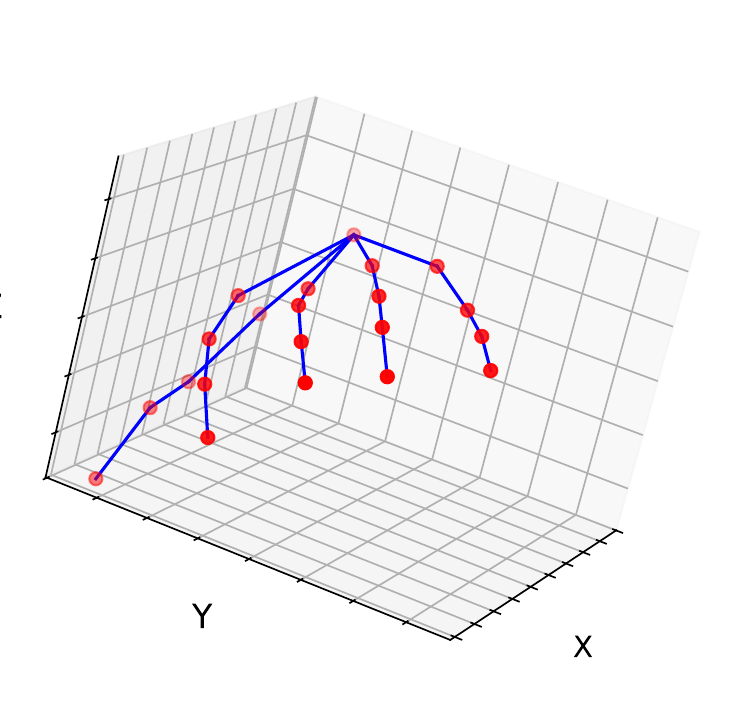}}
\caption{Qualitative comparison of 3D hand reconstructions obtained (\textbf{a}) with nearest-frame synchronization and (\textbf{b}) with sub-frame synchronization (triangulation from GoPro 3 and 4), for the scene shown in Figure~\ref{fig:hand_scene} (left hand). The nearest-frame approach produces noticeable discrepancies in hand shape, whereas sub-frame synchronization yields a reconstruction consistent with the true hand configuration.\label{fig:hand_comparison}}
\end{figure}

Finally, we present a qualitative comparison with and without sub-frame synchronization in Figure~\ref{fig:hand_comparison}(a) and (b), respectively, associated with the scene shown at the top of Figure~\ref{fig:master_comparison}(a). We observe a clear discrepancy in hand shape under the nearest-frame strategy, whereas the sub-frame synchronization case produces a reconstruction consistent with the actual hand shape.

\begin{table}[H]
\caption{Mean triangulation error (pixels) of 3D hand joints triangulated using four calibrated GoPros across six selected moments ($T_1$–$T_6$) where the full hand is visible and in motion. Results are reported for both nearest-frame synchronization and sub-frame synchronization, demonstrating the benefit of sub-frame accurate synchronization.\label{tab:mre_summary2}}
\centering
\begin{tabular}{l|cccccc|c}
\toprule
\textbf{Method} & $\mathbf{T_1}$ & $\mathbf{T_2}$ & $\mathbf{T_3}$ & $\mathbf{T_4}$ & $\mathbf{T_5}$ & $\mathbf{T_6}$ & \textbf{Mean $\pm$ Std} \\
\midrule
Nearest frame & 4.51 & 5.16 & 9.78 & 5.11 & 20.85 & 7.36 & $\boldsymbol{8.80 \pm 6.22}$ \\
Sub-frame interpolation & 3.85 & 3.20 & 5.67 & 3.15 & 5.08 & 5.44 & $\boldsymbol{4.40 \pm 1.14}$ \\
\end{tabular}
\end{table}

%% Comparing euclidian distance between triangulations using different gopros for interpolated and uninterpolated 

% -------------------------------------------------------------------------

\subsection{Experiment 5: 3D reconstruction}
In this experiment, we visually demonstrate the benefits of sub-frame accurate synchronization for multi-view 3D reconstruction. We combined our synchronization method with a state-of-the-art video interpolation technique to generate temporally aligned frames across all viewpoints, thereby significantly enhancing the quality of 3D reconstruction of dynamic scenes, particularly at lower frame rates.

\begin{figure}
\centering
\subfloat[\centering]{\includegraphics[width=5cm]{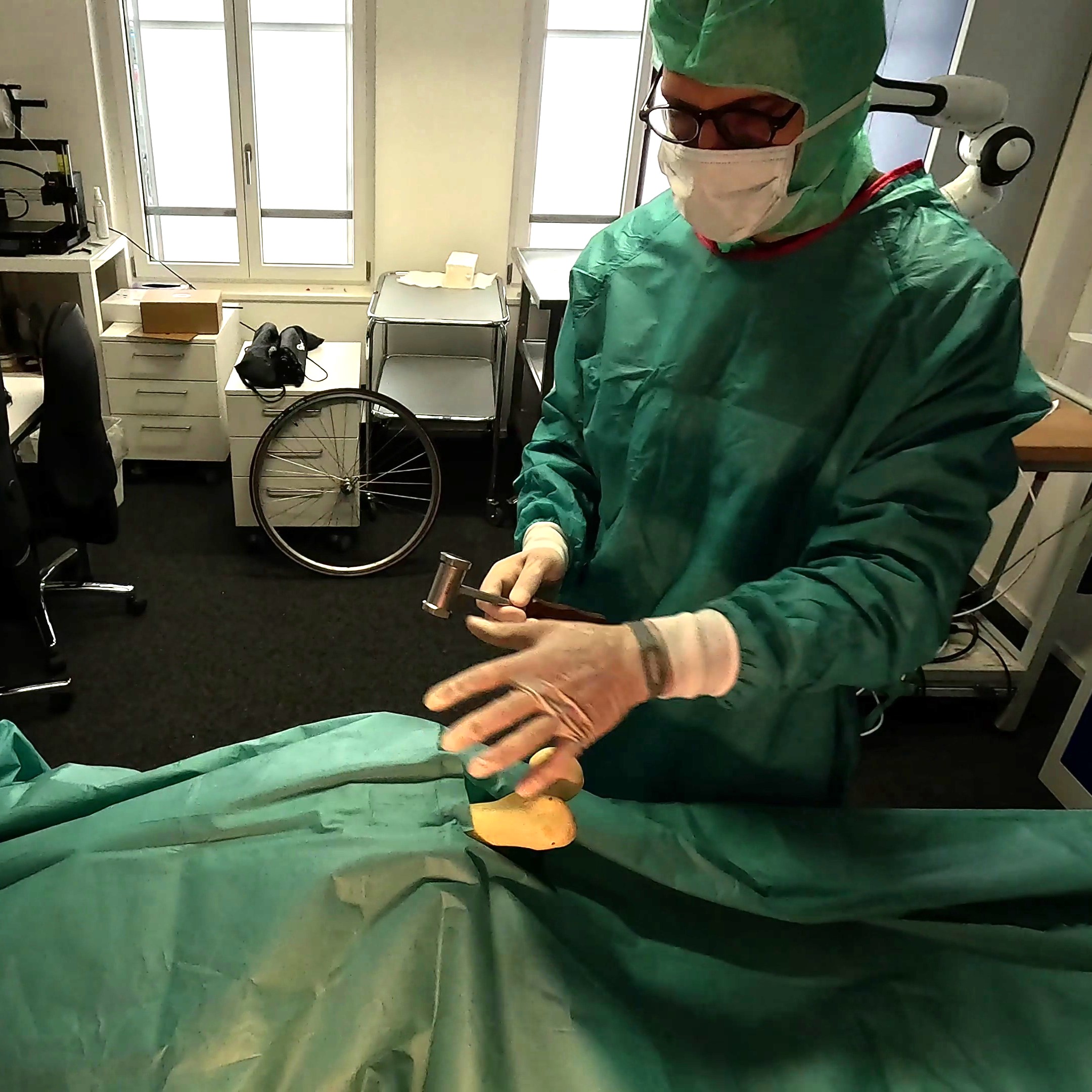}\label{fig:hand_scene}}
\hspace{0.5cm}
\subfloat[\centering]{\includegraphics[width=5cm]{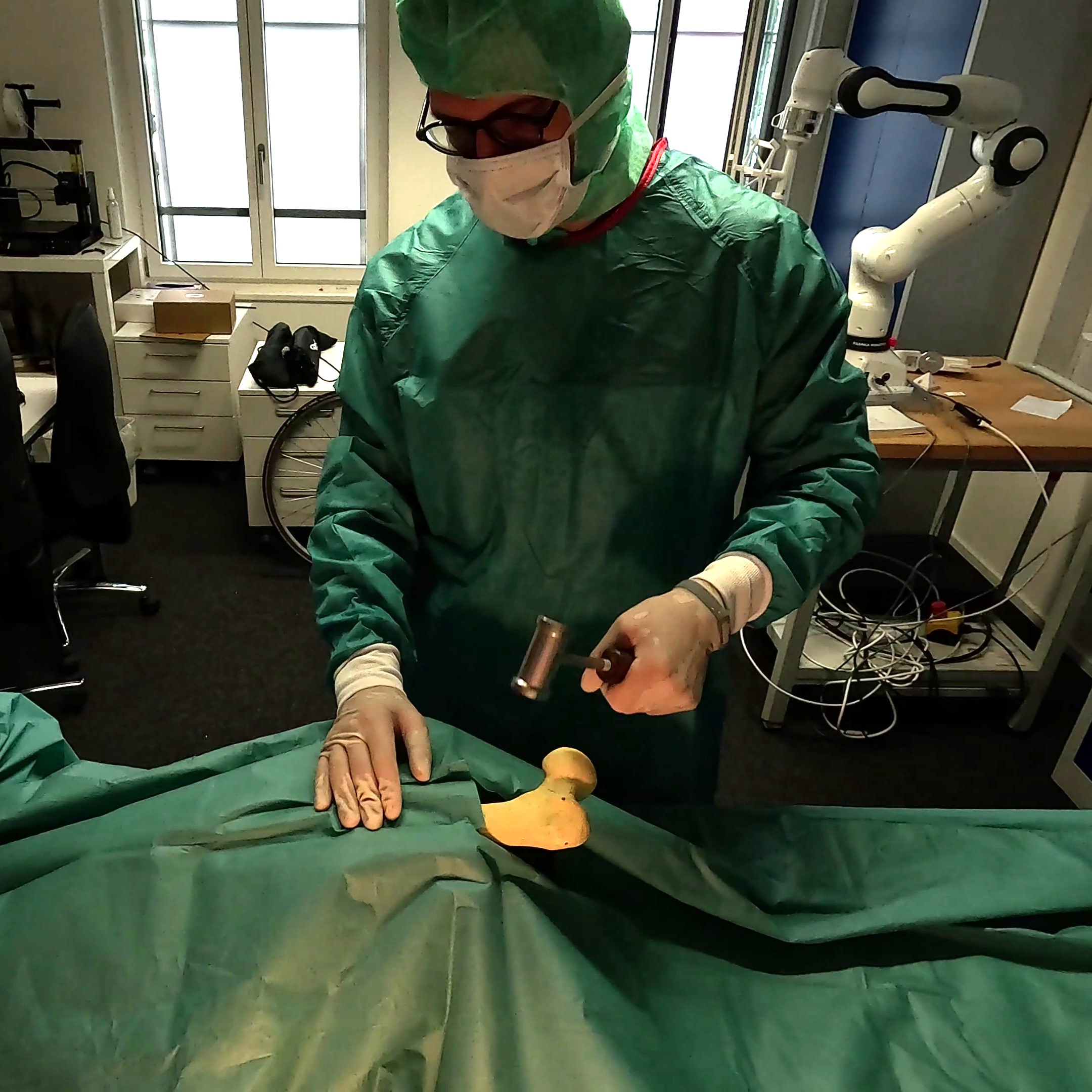}}\\
\subfloat[\centering]{\magnifyimage{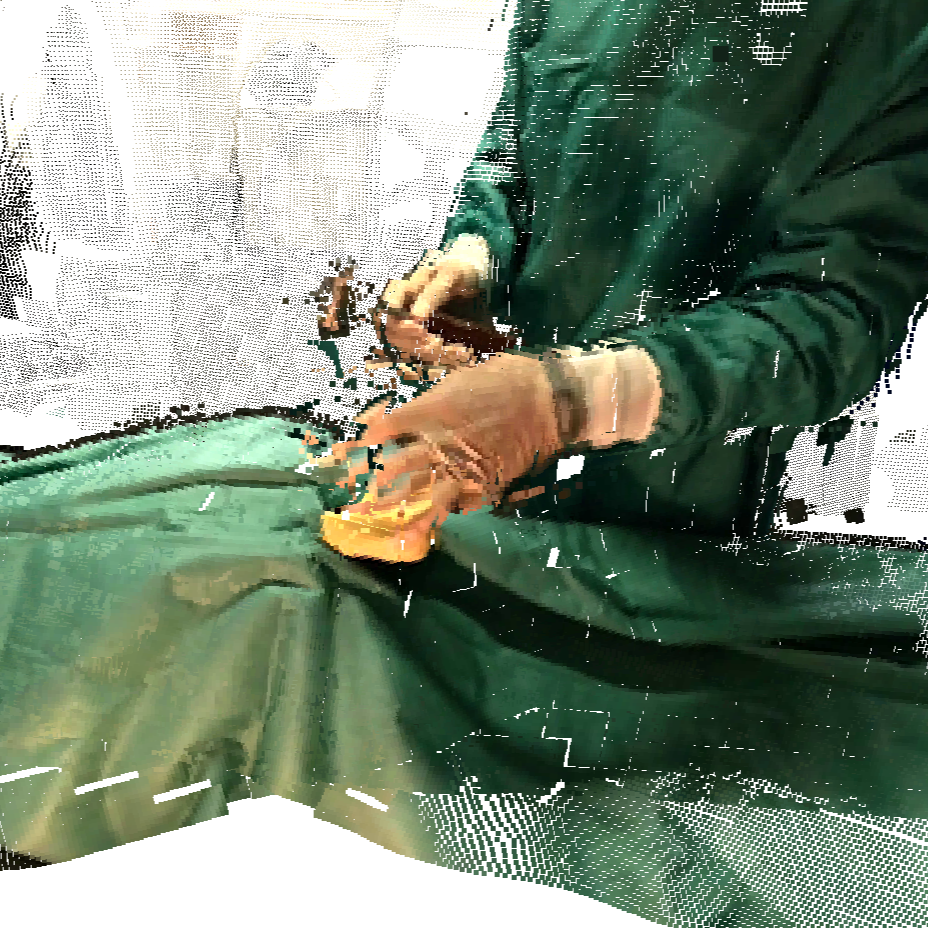}{-0.4}{0.1}{0.5}{1.25}{-1.25}}
\hspace{0.5cm}
\subfloat[\centering]{\magnifyimage{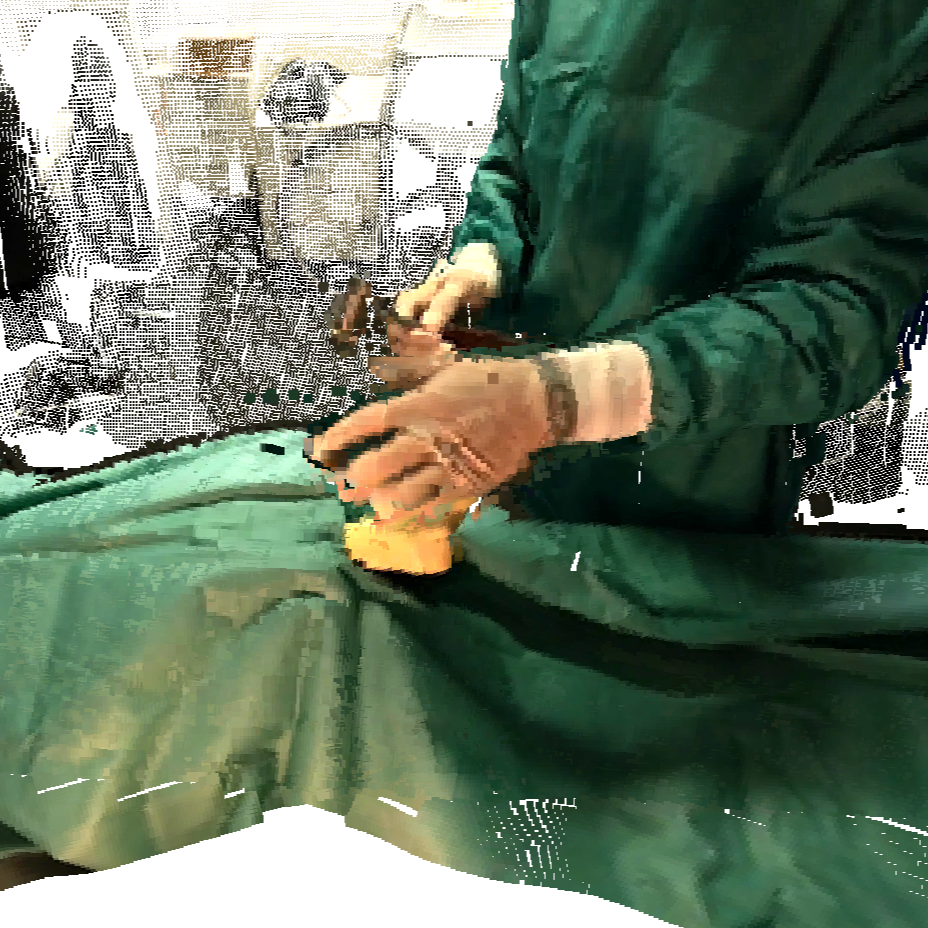}{-0.4}{0.1}{0.5}{1.25}{-1.25}}\\
\subfloat[\centering]{\magnifyimage{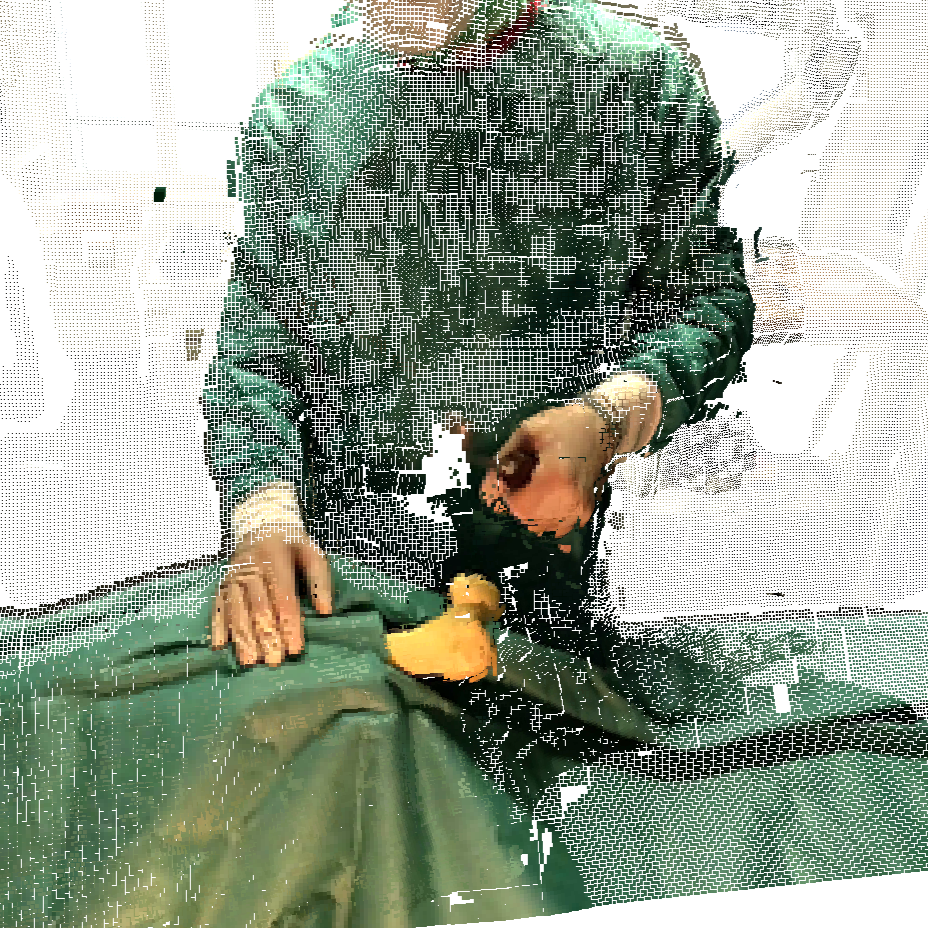}{0}{0}{0.5}{1.25}{-1.25}}
\hspace{0.5cm}
\subfloat[\centering]{\magnifyimage{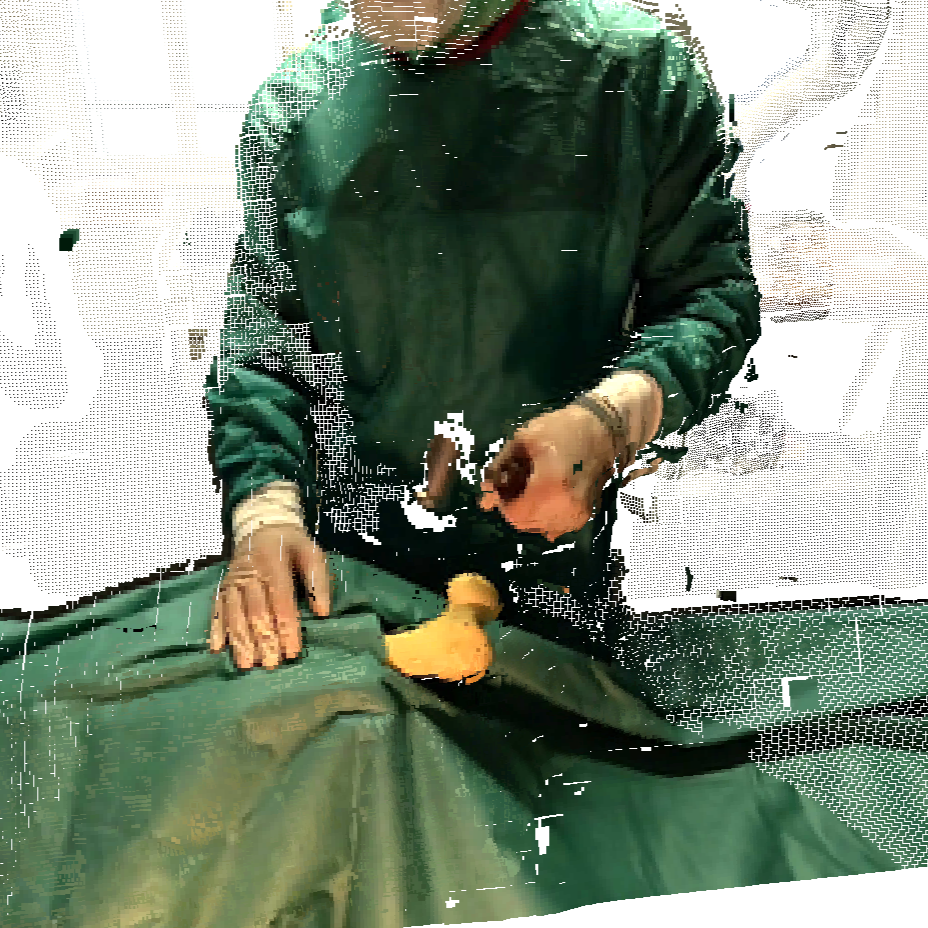}{0}{0}{0.5}{1.25}{-1.25}}
\caption{Comparison of 3D reconstruction results without and with sub-frame synchronization. Top: reference real images corresponding to the reconstructions shown in the middle panels (\textbf{a}) and the bottom panels (\textbf{b}), respectively. Middle: The left hand using a nearest-frame synchronization strategy presents large reconstruction artifacts (\textbf{c}) while being successfully reconstructed using the proposed method that allows for sub-frame synchronization (\textbf{d}). Bottom: 3D reconstruction based on the  nearest frame-based synchronization strategy fails in reconstructing the hammer (\textbf{e}) which is mostly reconstructed using sub-frame interpolation (\textbf{f}) (see image center). Panels (\textbf{c})–(\textbf{d}) use footage from GoPro~1, GoPro~3, and GoPro~4; panels (\textbf{e})–(\textbf{f}) use GoPro~2 and GoPro~3.\label{fig:master_comparison}}.
\end{figure}

\paragraph{Experimental setup}
We reused the recordings from Experiment~4 and selected two moments with significant motion. The method of Jin et al.~\cite{jin2025unifiedarbitrarytimevideoframe} was applied to generate temporally aligned frames across all viewpoints. MASt3R~\cite{leroy2024groundingimagematching3d} was then performed on both the interpolated and non-interpolated frames. The reconstructions are shown in Figure~\ref{fig:master_comparison}.

\paragraph{Results} For both selected moments, the interpolated frames produce noticeably higher-quality reconstructions with fewer artifacts compared to the nearest-frame strategy. We conclude that sub-frame interpolation using our synchronization method provides a practical and effective alternative to hardware synchronization for scenes involving fast-moving objects.

% -------------------------------------------------------------------------

\subsection{Large-scale data collections}
In addition to controlled experiments, our method has been successfully deployed in multiple surgical data collections, recording several hours of surgery footage using more than 25 cameras. These collections featured a highly heterogeneous camera array, including near-field and far-field GoPro cameras operating in linear lens mode, head-mounted Aria glasses capturing a wide-angle view, Canon CR-N300 capturing close-up views, different iPhone models, Aria glasses, and a high-end infrared surgical navigation system (Atracsys FusionTrack 500) for tool tracking. The devices exhibited significant variations in shutter speeds, frame rates, sensor types, and price points.

This diversity posed considerable synchronization challenges that are not addressed by existing methods. The successful synchronization of all video streams highlights the flexibility and robustness of our approach in complex, real-world environments. Synchronization was manually confirmed by visually comparing frames with significant motion across all cameras.

\begin{figure}[H]
\centering
\includegraphics[width=10cm]{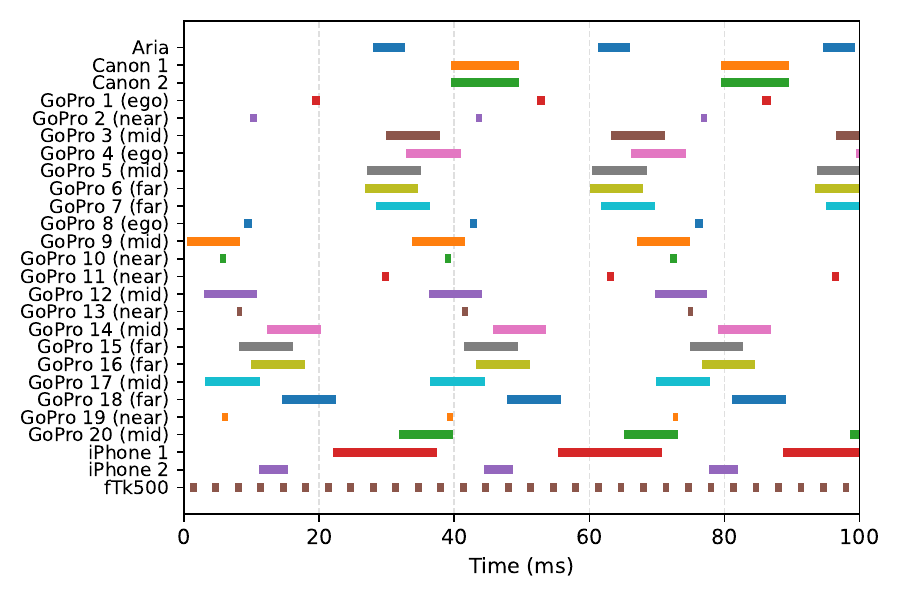}
\caption{Measured frame timing across 25 unsynchronized cameras. The variation in bar lengths indicates that cameras use different exposure settings. Exposure windows for near-field and egocentric cameras are markedly shorter than those of the far-field cameras due to the high intensity of the surgical lamp. The Atracsys fusionTrack 500 (fTk500), a stereo infrared camera, operates at a much higher frame rate than the RGB cameras.\label{fig:exposure_plot}}
\end{figure}

Figure~\ref{fig:exposure_plot} illustrates the relative exposure times of 25 unsynchronized cameras used simultaneously in a single recording. Our method accurately determines the exact start and end of each exposure. Note that the exposure durations differ across cameras, as their configurations were adjusted according to the amount of light within each field of view. Specifically, the \textit{near} and \textit{ego} groups have considerably shorter exposure times.

To validate our synchronization method, each video was visually inspected at a minimum of two checkpoints. We extracted frames from the aligned video streams and checked for light switch events, where the surgical lamp or the room light was turned off and on at the beginning and end of each recording.

In cases where a camera stopped recording early (e.g., egocentric cameras worn by surgeons leaving the operating room), the second light-switch event was not available. In those situations, we relied on distinct, fast movements or other visible markers to verify the synchronization at the second checkpoint. Synchronization between the IR and RGB streams was validated separately in Experiment 3.

Out of 73 total recordings, only one instance failed synchronization. This was associated with the Aria glasses worn by the lead surgeon. The failure was not due to our method, but rather to frame drops in the recordings from the Aria glasses.

\section{Conclusions}
We have presented a novel tandem hardware and software solution for millisecond-accurate temporal multi-view video synchronization that is compatible with a wide range of camera systems and has been successfully validated in a broad range of scenarios. 

Through extensive experiments, we have demonstrated that for synchronizing two GoPro action cameras, our method outperforms all existing approaches. Moreover, for systems including infrared cameras, our approach remains effective. 

We have also demonstrated that sub-frame accurate synchronization can be achieved without hardware synchronization and can enhance the performance of many downstream computer vision tasks through frame interpolation. This allows operation at lower frame rates while still recovering precise information, reducing storage and computational requirements, and enabling more efficient large-scale data collection and analysis.

Finally, our method has been successfully tested outside of the lab with 25+ cameras, demonstrating its robustness and real-world applicability. Together, these results confirm that our approach provides accurate, reliable, and broadly applicable sub-frame synchronization across diverse camera systems and practical scenarios.

% =========================================================================

\section{Future Work}
In this work, we evaluated our method using cameras operating at up to 60~fps. An interesting direction for future research is to extend the evaluation to high-speed cameras (1000+~fps), where temporal precision becomes even more critical for frame-perfect synchronization. Additionally, the computer vision pipeline could be enhanced to compensate for rolling shutter artifacts. Another promising direction is the investigation of real-time processing methods to enable synchronization of live video streams, rather than performing offline synchronization during post-production.

\vspace{2em}

% admin stuff
\paragraph{Author Contributions} Conceptualization, L.C.; methodology, J.M. and L.C.; hardware, J.M. and F.G.; software, J.M.; investigation, J.M., L.C. and J.W.; writing---original draft preparation, J.M.; writing---review and editing, L.C.; supervision, L.C.; project administration, M.P., P.F.; funding acquisition, P.F. All authors have read and agreed to the published version of the manuscript.

\paragraph{Funding} This work has been supported by the OR-X - a swiss national research infrastructure for translational surgery - and associated funding by the University of Zurich and University Hospital Balgrist, and the Innosuisse Flagship project PROFICIENCY No. PFFS-21-19.

\paragraph{Data Availability Statement} All source code used in this study is openly available in a public GitHub repository at \url{https://github.com/jaromeyer/RocSync}. The version used for the experiments corresponds to commit `5d82ddc`. The datasets generated and analyzed during the current study are available from the authors on reasonable request.

\paragraph{Acknowledgments} The authors would like to thank Alan Magdaleno and Max Krähenmann for their help in evaluating the experiments.
We also thank Sergey Prokudin and Jonas Hein for sharing data for the large-scale data collections.
During the preparation of this manuscript, the author(s) used ChatGPT for the purposes of correcting grammar and improving text flow. The authors have reviewed and edited the output and take full responsibility for the content of this publication.

% appendix
\clearpage
\appendix
\section[\appendixname~\thesection]{fusionTrack Pipeline}
\label{sec:ftk_pipeline}
The Atracsys fusionTrack 500 system required a modified processing pipeline compared to the one presented in Section~\ref{subsec:software}. Unlike standard cameras that output images, this system records sequences of 3D positions and orientations corresponding to detected fiducials and markers. Here, a \textit{fiducial} refers to a single reflective sphere or infrared LED, while a \textit{marker} denotes a predefined geometric arrangement composed of three or more fiducials. In our setup, the corner LEDs of our LED clock were registered as a marker. The fusionTrack system thus directly provides both the 3D position and orientation of this marker. The timestamp-encoding LEDs (i.e., counter and ring) are detected as individual fiducials, and the 3D positions of these fiducials are stored accordingly.

\begin{enumerate}
    \item \textbf{Detect marker frames:} Identify all frames in which the RocSync marker is successfully detected.
    \item \textbf{Transform fiducials to the local frame:} Transform all detected fiducials into the local coordinate frame defined by the marker.
    \item \textbf{Filter fiducials:} Retain only fiducials that lie close to the plane of the LED clock device.
    \item \textbf{Decode LEDs:} For each LED, check whether a nearby fiducial was detected; if so, interpret that LED as illuminated.
    \item \textbf{Estimate offset and drift:} Apply RANSAC-based linear regression to the decoded timestamps across multiple frames to compute a global offset and drift.
\end{enumerate}

\clearpage
\pagebreak

\section[\appendixname~\thesection]{Supplementary Figures}
\begin{figure}[htbp]
    \centering
    \includegraphics[width=7cm]{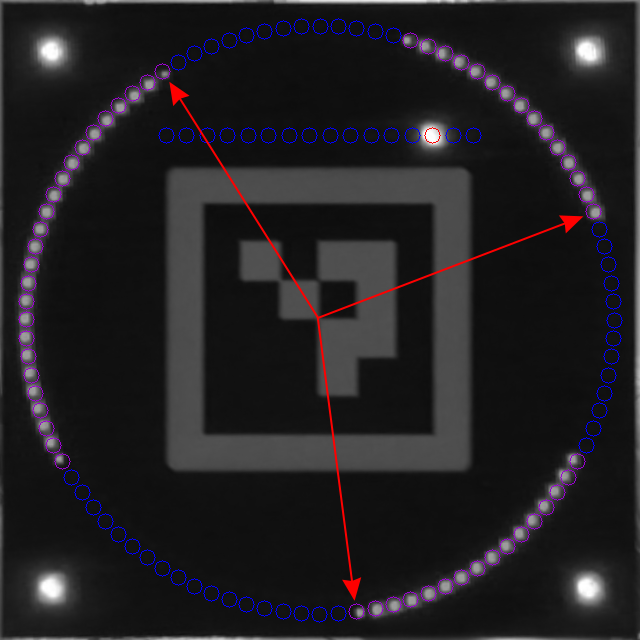}
    \caption{Three consecutive frames overlaid to demonstrate rolling shutter artifacts (from Experiment 1 with Azure Kinect DK). The beginnings of the three sectors (marked by the arrows) are unevenly spaced, and the sectors themselves differ in length.}
    \label{fig:kinect_rollingshutter}
\end{figure}

\bibliographystyle{unsrt}
\bibliography{bibliography}

\end{document}